# TomoSLAM: Factor Graph Optimization for Rotation Angle Refinement in Microtomography

Mark Griguletskii[2], Mikhail O. Chekanov[1,2], Oleg S. Shipitko[1,2], Ingacheva Anastasia[1,3], Vladislav Kibalov[1,2]

[1] Institute for Information Transmission Problems – IITP RAS, Bol'shoy Karetnyy Pereulok 19, Moscow, Russia, 127051.
[2] Evocargo, Moscow, Russia
[3] Smart Engines Service LLC, Moscow, Russia

## ABSTRACT

In computed tomography (CT), the relative trajectories of a sample, a detector, and a signal source are traditionally considered to be known, since they are caused by the intentional preprogrammed movement of the instrument parts. However, due to the mechanical backlashes, rotation sensor measurement errors, thermal deformations real trajectory differs from desired ones. This negatively affects the resulting quality of tomographic reconstruction. Neither the calibration nor preliminary adjustments of the device completely eliminates the inaccuracy of the trajectory but significantly increase the cost of instrument maintenance.

A number of approaches to this problem are based on an automatic refinement of the source and sensor position estimate relative to the sample for each projection (at each time step) during the reconstruction process. A similar problem of position refinement while observing different images of an object from different angles is well known in robotics (particularly, in mobile robots and self-driving vehicles) and is called Simultaneous Localization And Mapping (SLAM). The scientific novelty of this work is to consider the problem of trajectory refinement in microtomography as a SLAM problem. This is achieved by extracting Speeded Up Robust Features (SURF) features from X-ray projections, filtering matches with Random Sample Consensus (RANSAC), calculating angles between projections, and using them in factor graph in combination with stepper motor control signals in order to refine rotation angles.

## 1. INTRODUCTION AND RELATED WORK

Computed tomography is used in various fields of human activity: medicine [1], non-destructive testing in production, precision metrology [2] and others. Reconstruction algorithms rely on the geometric accuracy of the instrument and the precisely known relative angle between a signal source and an object being scanned [3]. In this regard, a significant amount of effort is regularly spent on the calibration of tomography systems [4]. Geometric calibration is usually a labor-intensive and expensive process that requires the involvement of qualified specialists. Moreover, the quality of the calibration degrades over time, and the geometric errors inherent in a calibrated system can still have a negative impact on the quality of the reconstruction.

The problem of geometric errors correction has been widely studied [5]. There are two main types of geometric calibration:

- calibration based on a reference object - an object with a known geometry, which makes it possible to evaluate and correct the influence of errors by methods of projective geometry [6];

- calibration based on reference measuring devices that allow direct measurement of coordinate inaccuracies in the tomographic scheme [7].

Different methods are developed to eliminate a certain type of errors (scaling errors [8], errors in measuring rotational motion [9], etc.). The calibration procedures using these methods do not allow one to take into account the factors arising from the difference in the conditions of calibration experiments and experiments with real objects. These factors include thermal deformations resulting from a significantly longer duration of a real tomographic experiment as compared to an experiment conducted for calibration, as well as errors caused by a change in the geometry of the test sample or its movement under its own weight during the measurement.

An alternative approach to refining the result of tomographic reconstruction is the so-called on-line calibration. On-line methods estimate and correct geometric errors during a tomographic experiment [10, 11, 12]. There is ongoing active research in this area. Since the developed methods are computationally expensive, in practice, preference is still given to the methods of calibration against a reference object described earlier [13]. In addition, on-line calibration is considered one of the blocking technologies in the construction of ultrasound tomography systems [14].

According to the works published in the field of CT, there is a gradual transition from methods with a priori known position and shape of the object under calibration [15] to methods with a simultaneous refinement of the calibration phantom and parameters [11]. A similar transition occurred in the field of mobile robots localization. Research focus and methods changed from the localization with a known map [16] to simultaneous localization and mapping [17]. In robotics, this transition took place around 2007 [18], and since then many solutions to this problem have been proposed, with the main progress being observed in solving the SLAM problem in real-time as the robot moves. The task of geometric errors correction in tomography can be directly considered as the task of simultaneous refining the position of the detector ("L", robot's localization in SLAM terminology) and the object reconstruction ("M", mapping in SLAM terminology). Thus, there is reason to believe that adaptation of the methods from robotics developed for solving the SLAM problem can significantly accelerate progress in solving the problem of decreasing geometric errors in the computed tomography. In this work, we focus on the localization part (rotation angle refinement) and consider the mapping part (tomographic reconstruction methods) being solved by existing methods.

The graph-based formulation of SLAM problem was initiated by Lu and Milios in 1997 [19]. A significant number of modifications to the original algorithm were developed over the following decades. State-of-the-art approaches fuse information from different sensors [20] as well as use Deep Neural Networks to extract features/landmarks to reconstruct objects and map [21]. Moreover, in [22] a modification of the algorithm is proposed, which makes it possible to take into account the dynamic changes of an observed object. While the re-observing same area of the object, the algorithm allows to identify the presence of changes in the structure of the object and take them into account, dynamically updating the connectivity graph. One of the weak points of SLAM algorithms is its sensitivity to false associations - incorrectly identified dependencies between two observations of the object. False data associations can lead to a significant deterioration in terms of mapping and trajectory quality. To deal with false associations it is proposed to develop more stable descriptors of visual features [23] as well as filter incorrect matches [24]. An alternative approach described in [25] is the development of graph optimization algorithms that are resistant to association errors. The proposed algorithms make it possible to determine false associations in the process of graph optimization and neutralize their influence on the resulting trajectory.

In this work, we propose a method of adapting SLAM technology to decrease accumulating rotation errors for computed tomography based on factor graph construction, where motor, rotating the instrument, and computer vision rotation angle estimation method are used as factors. We also experimentally estimate the best parameters of our approach and evaluate the limitations of the method. For experimental evaluation, a dataset of 360 X-ray images observing micro-SD card has been collected. Although, in this work we only consider simple motion model with only rotation error, the proposed method can be extended to solve the full 3D pose estimation as it successfully showed in 3D graph SLAM approaches in the field of robotics [22].

## 2. PROBLEM STATEMENT

In our experiments, an X-ray camera (X-ray source and detector) is fixed in space whereas an object rotates around its axis of rotation with an assumption of zero axial displacement. The object is rotated by a stepper motor. This is equivalent to the model with the stationary object and rotating camera. Without loss of generality further, we consider this setup. Therefore stepper motor rotation is directly translated to the object. Besides zero axial displacement, as an assumption, there is no any backlashes except rotational. Let $I_1, I_2, I_3$ are the X-ray images obtained from the detector in three different positions lying on a circular path. The rotation axis is parallel to the Y-axis of the detector and is stationary for the images, as Figure 1 illustrates. Let $\Delta_1$ and $\Delta_2$ are relative rotation angles of detector between poses $\theta_1, \theta_2, \theta_3$ in which $I_1, I_2, I_3$ are taken respectively. Their rotation angles are subject to measurement error and need to be refined. We assume that x coordinate of the rotation axis is known, equals $0$, and is collinear to the rotating shaft of the stepper motor. Given a sequence of images $I_i, i = \overline{1, N}$ obtained from detector, equations (2) allow to calculate rotation angles between frames. Section 3 explains how we choose correct features matches and estimate angles based on them. Section 4 describes a fusion process of these different measurements between the same frames which lead to refinement of rotation angles through the whole trajectory.

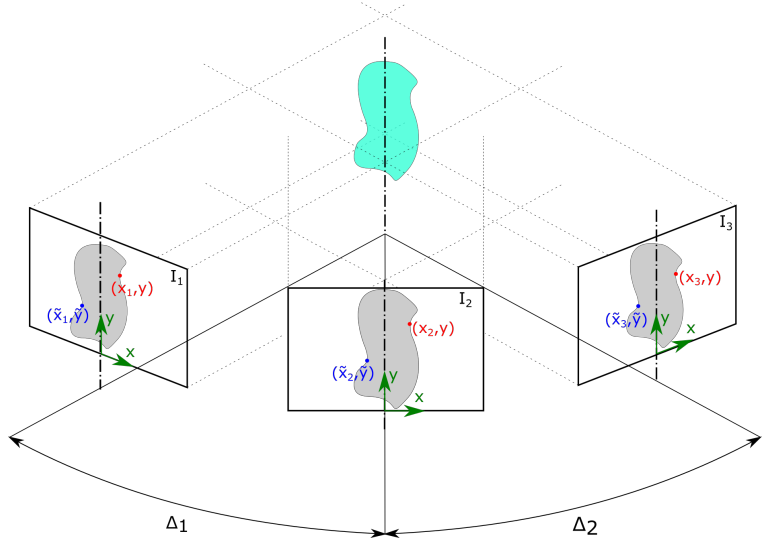

Figure 1. Model of a tomographic setup

Each object point projection in the image has one degree of freedom – $x$ coordinate in pixels which is described as follows:
$$x(\alpha) = a \cos(\alpha + \phi_0),$$
where $\alpha$ is detector rotation angle, $a$ and $\phi_0$ are the parameters specific for each point. One can show that having poses $x_i, \tilde{x}_i, i = \overline{1,3}$ of two object's points projection for three images it is possible to formulate a system of equations by combining these poses:

$$\begin{cases} x_1 = a\cos(\phi_0), \\ x_2 = a\cos(\Delta_1 + \phi_0), \\ x_3 = a\cos(\Delta_1 + \Delta_2 + \phi_0), \\ \tilde{x}_1 = b\cos(\psi_0), \\ \tilde{x}_2 = b\cos(\Delta_1 + \psi_0), \\ \tilde{x}_3 = b\cos(\Delta_1 + \Delta_2 + \psi_0), \end{cases} \rightarrow \begin{cases} \cos\Delta_1 = \dfrac{(1-k^2)x_1^2 - (x_3 - kx_2)^2}{2kx_1(x_3 - kx_2)}, \\ \sin(\Delta_1 + \Delta_2) = k\sin\Delta_1, \\ k = \dfrac{x_3\tilde{x}_0 - x_0\tilde{x}_3}{x_2\tilde{x}_0 - x_0\tilde{x}_2}. \end{cases} \quad (1)$$

If we set the limitations for $\Delta_i$ and $x_i, \tilde{x}_i$, such that $k$ exists and is limited, $\Delta_i$ is positive and $\Delta_1 + \Delta_2 < \frac{\pi}{2}$ the system has the only solution:

$$\begin{cases} \Delta_1 = \arccos \dfrac{(1-k^2)x_1^2 - (x_3 - kx_2)^2}{2kx_1(x_3 - kx_2)}, \\ \Delta_2 = \arcsin k\sin\Delta_1 - \Delta_1, \\ k = \dfrac{x_3\tilde{x}_0 - x_0\tilde{x}_3}{x_2\tilde{x}_0 - x_0\tilde{x}_2}. \end{cases} \quad (2)$$

## 3. COMPUTER VISION BASED ROTATION ANGLE ESTIMATION

As there might exist more than 2 tracked features (more than one pair) in the triplet of images $I_1, I_2, I_3$ (see Figure 2) we introduce an algorithm to estimate angles between frames taking into account multiple matched points pairs. SURF [26] keypoint detector and descriptor are used for tracking points on a sequence of images. To match keypoints on each image and get its tracks we find nearest by L2-metric keypoints descriptors for $I_1 - I_2, I_2 - I_1, I_2 - I_3, I_3 - I_2$ image pairs. We consider matched keypoints to be a good track if (1) a pair of matched keypoints for $I_1 - I_2$ images is the same pair for $I_2 - I_1$, the same is true for $I_2, I_3$ image, (2) matched keypoints' y-axis shift is less than 2 pixels.

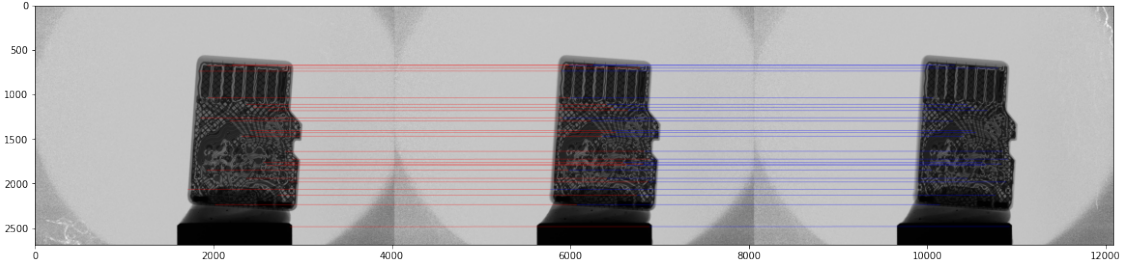

Figure 2. Example of features tracks in a triplet of images with 10 degrees step size from collected dataset

With the obtained points' tracks, it is possible to evaluate rotation angles for each pair of tracks, however different pairs give different angle estimations due to noise of keypoints detection and outliers (incorrectly matched keypoints). To achieve angle estimation most consistent with inliers (correctly matched keypoints) and robust to the presence of outliers we use RANSAC-based algorithm [27]. On each iteration of the algorithm with a randomly picked pair of tracks, we calculate $k$ with equation (2). Next, we go through all tracks and substitute each track to equation for $k$ (2) along with first or second track from the picked pair to calculate $\tilde{k}_1, \tilde{k}_2$ respectively. If both $(k - \tilde{k}_1)^2$ and $(k - \tilde{k}_2)^2$ values are less than fixed threshold, we consider track is an inlier. After all iterations are finished we take track pair with the highest number of inliers to compute $\Delta_1$ and $\Delta_2$. Further, we will refer to this algorithm as CV based angle estimation. Number of algorithm iterations $M$ is calculated as follows:

$$M = \frac{1-p}{1-q^2},$$

where $q$ is a probability to pick a reliable track, $p$ is a desired probability that algorithm calculated reliable $k$ at least once.

## 4. ROTATION ANGLE REFINEMENT WITH FACTOR GRAPH

In robotics, the factor graph approach [28] maximizes the likelihood of robot's positions X and observed landmarks M given observation Z, control signals U which are normally distributed. With this assumption, it becomes possible to interpret this problem as least-squares minimization and efficiently solve it. In this work we assume that controls signals U for stepper motor as well as CV based angle estimations are also normally distributed. As we do not reconstruct a map, equation (3) represents Bayesian inference for our problem with MAP (Maximum A Posterior):

$$\mathrm{X}^{MAP} = \underset{X}{\mathrm{argmax}}\, P(X|U^{CV}, U^{Enc}) = \underset{X}{\mathrm{argmax}}\, \frac{p(U^{CV}, U^{Enc}|X)P(X)}{P(U^{CV}, U^{Enc})}. \quad (3)$$

Figure 3 shows a part of a full circle camera trajectory. It depicts a graph with nodes $(X_1, X_2, .., X_n)$ and edges of different factors: $(CV_1, CV_2, .., CV_n), (Enc_1, Enc_2, .., Enc_n)$. CV-factor links two poses with an angle calculated by CV based angle estimator. A step size between images can be variable. Experiments to choose the best step size are described in Section 5. Meanwhile, Enc-factor links every 2 nodes sequentially based on control signals for the stepper motor. Then we use this step to link angles $\theta_i, \theta_j$ in the factor graph.

Equation (4) describes the PDF (Probability Density Function) of angles $\theta$ given measurements $u^{Enc}, u^{CV}$ and a likelihood to be maximized:

$$P(\theta) = p(\theta_0) \prod_{i=1}^{N} p\left(\theta_i | \theta_{i-1}, u_i^{Enc}\right) \prod_{j=1}^{M} p\left(\theta_j | \theta_{j-1}, u_j^{CV}\right), \quad (4)$$

$\theta \in R^1$ - camera rotation angle in 1D.

With known prior $P(\theta_0)$ one can maximize the likelihood by minimizing its minus logarithm as follows:

$$\theta = \underset{\theta}{\operatorname{argmax}} P(\theta) = \underset{\theta}{\operatorname{argmin}} \{-log(P(\theta))\}$$

$$= \underset{\delta\theta_{i,j}}{\operatorname{argmin}} \sum_{i=1}^{N} ||W_{Enc}(\delta\theta_{i-1} - \delta\theta_i) - W_{Enc}u_i^{Enc}||_2^2 + \sum_{j=1}^{M} ||W_{CV}(\delta\theta_{j-1} - \delta\theta_j) - W_{CV}u_j^{CV}||_2^2, \quad (5)$$

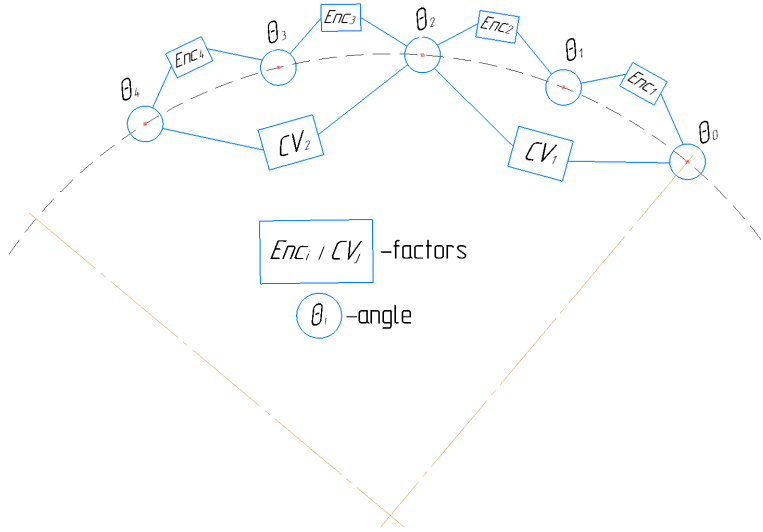

Figure 3. Factor graph representation of the experiment

where $W_{Enc/CV} = \frac{1}{\sigma_{Enc/CV}}$, $\sigma_{Enc/CV}$ – standard deviation of stepper motor controls signals and CV based angle estimator measurement noise.

Equation (6) is equivalent to over-constrained system of linear equations which could be efficiently solved with pseudo-inverse of $A^T A$ with the help of Cholesky decomposition [29]:

$$||A\delta - b||_2^2 \to min, \quad \to \quad \theta_i = \theta_i + \delta\theta_i. \quad (6)$$

## 5. EXPERIMENTAL EVALUATION

To test the proposed algorithm, a set of tomographic projections was collected from the X-ray microtomography device developed and operating at the Federal Research Center for Crystallography and Photonics of Russian Academy of Sciences [30]. Projections of a rotating object were captured with a step of 1 degree in the range of $0 - 360$ degrees (1 image per degree). The standard deviation of stepper motor positioning equals $\pm 0.05$ degrees.

### 5.1 Best Step Size for CV Based Angle Estimator

In this section, we describe an experiment for determining the best step size (relative angle) between images. The proposed CV based angle estimator has been launched with different step sizes on the whole set of 360 images. Figure 4 depicts median, first and third quartiles for every step size. It is worth noting that the amount of angles for statistics calculation is different for every step and the underestimation of variance might be present for large step sizes due to the limitation of 360 images per circle in the used dataset. Angles of 4/5/7/10-degrees have the least error distributions among others. However, a 7-degrees angle has bigger a median angle estimation bias compared to other step sizes. Thus, 4/5/10-degrees angles are good candidates as CV-factor step size in the factor graph. Besides this set of angles bigger 20/30/40-degrees and smaller 1/2/3-degrees applicants have been tested. Unfortunately, the proposed CV angle estimation approach did not show qualitative results for such distances due to the lack of tracked features between frames.

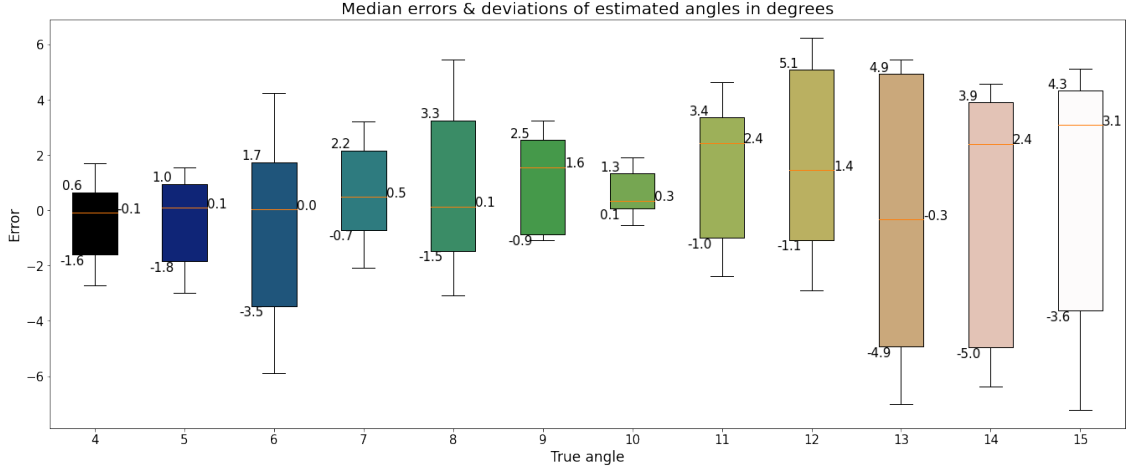

Figure 4. Median errors and quartiles between true and estimated angles by CV based algorithm with different steps

To illustrate the consistency of the assumption that measurements are normally distributed Figure 5 depicts the histograms of $\Delta_1$ and $\Delta_2$ angles between frames with 10-degrees step size.

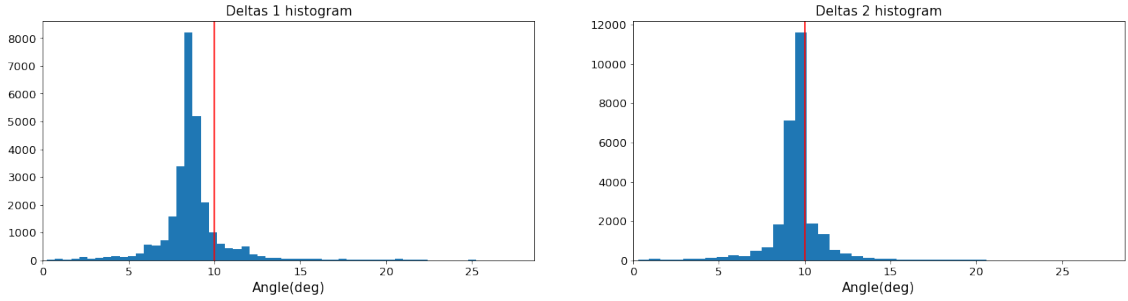

Figure 5. $\Delta_1$ and $\Delta_2$ histograms of CV based angle estimator for 10 degrees step size

### 5.2 Error Ratio Estimation With Different Parameters of Factor Graph

As the aim of the research is to show that accumulated rotation errors from stepper motor and mechanical backlashes might be reduced with the help of the factor graph approach with a combination of CV-factors it is meaningful to estimate the limitations of such a method. Figure 6 depicts the dependency of error ratio of root-mean-squared error (RMSE) of pure stepper motor angle estimation to graph based one from CV/Enc-factors standard deviations:

$$RMSE = \sqrt[2]{\frac{\sum_{i=1}^{N}\left(\theta_i - \widetilde{\theta}_i\right)^2}{N}}, \quad \rightarrow \quad Error\ ratio = \frac{RMSE_{\text{pure stepper motor trajectory}}}{RMSE_{\text{graph based trajectory}}} \quad (7)$$

where $\widetilde{\theta}_i$-reference rotation angle.

RMSE is estimated relative to a noise-free trajectory which is represented by 360 uniformly distributed positions of the camera. Pure stepper motor trajectory is generated by adding normally distributed $\mathcal{N}\left(0,\ 0.05^2\right)$ to every angle. Graph based trajectory is a set of rotation angles after solving the factor graph. For each set of parameters (CV step angle and stepper motor standard deviation), we performed $500$ graph optimizations to obtain averaged statistics.

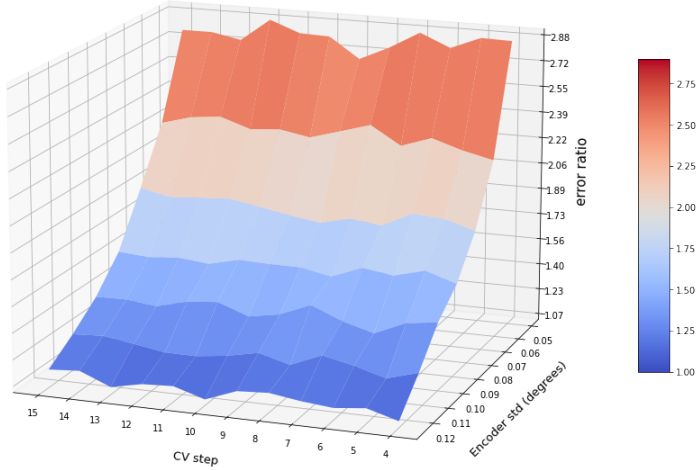

Figure 6. Dependence of error ratio from standard deviations of stepper motor control signals (Enc-factor) and CV based angle estimator (CV-factor)

The obtained results (see Figure 6) clearly show that RMSE of pure stepper motor trajectory might be $2.75$ times reduced using factor graph approach with CV-factors of the proposed CV based angle estimator with a 10-degrees angle step size. Additionally, a standard deviation of the stepper motor possibly could be increased by $2.4$ times from the initial $\pm 0.05$ degrees, and still, the same accuracy could be achieved. The lowest limit of the error ratio dependency equals $\approx 1$, which means that further stepper motor positioning noise increase leads to a quality decrease and there is no sense to use the proposed approach, as in this case it results in larger RMSE than pure stepper motor based estimation. It takes $0.053$ seconds to compute full 360-degree trajectory on the PC with Intel core i7 10850H CPU with the program written in Python and using 12 cores for parallel computing. Thus, the optimized trajectory can be calculated in real time as it calculated in many state-of-the-art SLAM approaches [31].

Figure 7 illustrates the difference in reconstructed SD card with the use of stepper motor (original) angles and refined ones obtained with the proposed algorithm. Reconstruction was performed with Smart Tomo Engine [32]. One can notice that the angles refinement led to less blurred reconstruction results. Also, more details on sample internal structure become evident after the refinement.

## 6. CONCLUSION

As a result of the research, we propose a method of accumulated rotation error reductions for the computed tomography problem. Additionally, we introduce a computer vision based algorithm for angle estimation between tomographic projections and evaluate its performance and limitations. As a result, factor graph representation of rotation angle estimation problem allows minimizing accumulated rotation error by solving over-constrained system of linear equations which leads to $2.75$ times RMSE reduction over pure stepper motor based estimation or alternatively keeps the same precision with $2.4$ times noisier stepper motor. This might reduce the price of stepper motor which potentially leads to cost reduction of tomographic equipment. As further research direction, computer vision angle estimation might be improved as common keypoints detection and description algorithms do not always perform well in X-ray images.

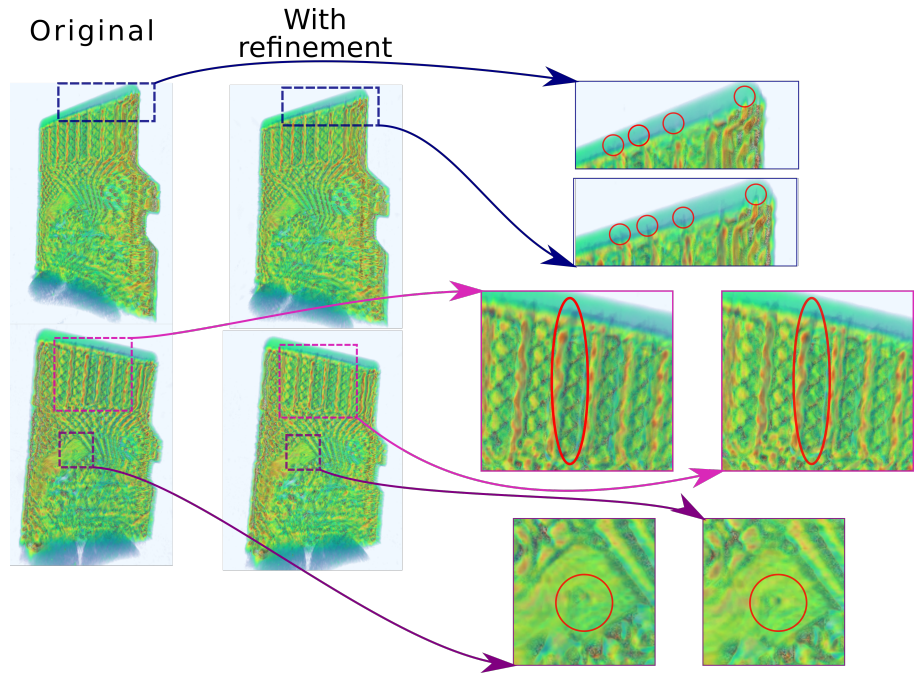

Figure 7. Tomographic reconstruction performed with Smart Tomo Engine [32] using original and refined angles.

## Acknowledgement

We express huge gratitude to Dmitry P. Nikolaev for consulting, review, and support as well as Alexey Buzmakov for the help in collecting X-ray dataset and performing the experiments with microtomograph. The reported study was partially funded by Russian Foundation for Basic Research, project numbers 18-29-26037 and 18-29-26028.